%% file: colm2025_conference.tex
\documentclass{article} 
\usepackage[final]{colm2025_conference}
\input{math_commands}

\usepackage{microtype}
\usepackage{hyperref}
\usepackage{url}
\usepackage{booktabs}
\usepackage{amsmath}
\usepackage{lineno}
\usepackage{graphicx}
\usepackage{multirow}
\definecolor{darkblue}{rgb}{0, 0, 0.5}
\hypersetup{colorlinks=true, citecolor=darkblue, linkcolor=darkblue, urlcolor=darkblue}

\title{Teach Old SAEs New Domain Tricks with Boosting
}

\author{Nikita Koriagin\thanks{Corresponding author: n.s.koryagin@tbank.ru}\\
T-Tech \\
\And
Yaroslav Aksenov \\
T-Tech\\
\And
Daniil Laptev \\
T-Tech \\
\AND
Gleb Gerasimov \\
HSE University \\
T-Tech \\
Moscow Institute of Physics and Technology \\
\AND
Nikita Balagansky \\
T-Tech \\
Moscow Institute of Physics and Technology \\
\And
Daniil Gavrilov\\
T-Tech
}

\begin{document}

\ifcolmsubmission
\linenumbers
\fi

\maketitle

\begin{abstract}
Sparse Autoencoders have emerged as powerful tools for interpreting the internal representations of Large Language Models, yet they often fail to capture domain-specific features not prevalent in their training corpora. This paper introduces a residual learning approach that addresses this feature blindness without requiring complete retraining. We propose training a secondary SAE specifically to model the reconstruction error of a pretrained SAE on domain-specific texts, effectively capturing features missed by the primary model. By summing the outputs of both models during inference, we demonstrate significant improvements in both LLM cross-entropy and explained variance metrics across multiple specialized domains. Our experiments show that this method efficiently incorporates new domain knowledge into existing SAEs while maintaining their performance on general tasks. This approach enables researchers to selectively enhance SAE interpretability for specific domains of interest, opening new possibilities for targeted mechanistic interpretability of LLMs.
\end{abstract}

\section{Introduction}
\label{sec:intro}
Large Language Models (LLMs) exhibit remarkable performance across numerous tasks, yet their internal mechanisms remain opaque. Mechanistic interpretability approaches, such as Sparse Autoencoders (SAEs), have emerged to disentangle LLM representations by mapping dense activations to higher-dimensional, sparse spaces \citep{bricken2023towards, cunningham2023sparse, gao2024scaling}. These sparse features often correspond to interpretable concepts, enabling researchers to isolate factors that govern model behaviors \citep{templeton2024scaling}.

However, SAEs can only capture frequently occurring features in their training data, leading to \textit{feature blindness} when encountering rare or domain-specific concepts that the LLM itself may have learned \citep{templeton2024scaling, muhamed2024decodingdarkmatterspecialized}. To address this issue, practitioners often retrain SAEs on domain-specific data \citep{saenanda, muhamed2024decodingdarkmatterspecialized}, but this process is computationally expensive and risks catastrophic forgetting. Fine-tuning new features individually may also create additional challenges if they fail to align with the existing feature set.

In this paper, we introduce a novel approach, \textit{SAE Boost}, that allows for selective enhancement of SAE capabilities without full retraining. Our method builds upon the existing SAE by training a supplementary model specifically designed to capture features missed by the original model when processing domain-specific texts. By modeling the reconstruction error on targeted domains, this supplementary SAE effectively learns complementary features that can be integrated with the base model's outputs during inference. 

Through empirical evaluations across multiple domains, we demonstrate improvements in both LLM cross-entropy and explained variance metrics when using the proposed method. SAE Boost enables researchers to selectively enhance their interpretability tools for domains of interest, facilitating more comprehensive analyses of LLM capabilities and limitations. By addressing the feature blindness problem, our work contributes to the broader goal of developing more robust and complete tools for the mechanistic interpretability of increasingly powerful language models.

\section{Related Work}
Sparse Autoencoders (SAEs) have emerged as powerful tools for the mechanistic interpretability of LLMs. Through sparse coding \citep{olshausen1996emergence, olshausen1997sparse}, they aim to decompose the dense activations of LLM layers into human-interpretable features. Recent work has demonstrated that SAEs can successfully extract interpretable features from LLMs \citep{bricken2023towards, cunningham2023sparse, gao2024scaling}, providing insights into how these models process and represent information. This approach builds on the broader field of dictionary learning, in which an overcomplete basis is learned to represent data sparsely \citep{mallat1993matching, olshausen1997sparse}. In the context of neural networks, it also connects to research on disentangled representations \citep{bengio2013deep, higgins2017beta} and feature visualization \citep{olah2017feature}.

Ensuring comprehensive feature coverage is a fundamental challenge in SAE training. \citet{leask2025sparseautoencoderscanonicalunits} showed that increasing dictionary size leads to two types of latents: novel latents that capture previously absent information, and reconstruction latents that refine existing features. They introduced ``SAE stitching,'' a method to integrate novel latents from larger SAEs into smaller ones, enhancing reconstruction without substantial model growth. \citet{templeton2024scaling} found that extremely large dictionaries (on the order of billions of features) are necessary to reliably capture rare concepts, underscoring the computational difficulty of addressing these infrequent yet critical features.

\section{Methodology}

\subsection{SAE Boost Architecture}

As noted in Section \ref{sec:intro}, fully fine-tuning an SAE with domain-specific data may degrade existing features. Moreover, tuning a small subset of newly initialized features (with arbitrary initialization) does not guarantee that these new features will precisely capture the missing domain features. To address this limitation, we propose a method designed to exactly cover missing features, which can be viewed as an error in the reconstruction of hidden states.

Our key insight is that, rather than retraining the entire SAE on domain-specific data, we can train a secondary SAE to model only the reconstruction errors of the pretrained SAE in the target domain. This method allows the model to learn additional features that the original SAE missed, without interfering with or disrupting existing representations. Our approach consists of two components: a pretrained SAE and a residual SAE, or SAE Boost. 

For a given input activation $\vx$ from an LLM layer, the pretrained SAE processes it as an autoencoder
\(
\hat{\vx} = \mW_{\text{dec}} \sigma(\mW_{\text{enc}} \vx + \vb_{\text{enc}}) + \vb_{\text{dec}},
\)
where $\mW_{\text{enc}} \in \mathbb{R}^{F \times d}$ and $\mW_{\text{dec}} \in \mathbb{R}^{d \times F}$ are the encoder and decoder weights, $\vb_{\text{enc}} \in \mathbb{R}^F$ and $\vb_{\text{dec}} \in \mathbb{R}^d$ are the encoder and decoder biases, and $\sigma$ is the sparsity-enforcing activation function (typically jumpReLU or batch\_topk) \citep{bricken2023towards}.

Instead of directly reconstructing $\vx$, the residual SAE learns the residual error $\ve = \vx - \hat{\vx}$ as
\(
\hat{\ve} = \mW_{\text{dec}}^{\text{res}} \sigma(\mW_{\text{enc}}^{\text{res}} \vx + \vb_{\text{enc}}^{\text{res}}),
\)
where $\mW_{\text{enc}}^{\text{res}}$ and $\mW_{\text{dec}}^{\text{res}}$ are the encoder and decoder weights of the residual SAE, and $\vb_{\text{enc}}^{\text{res}}$ is the encoder bias. We omit the decoder bias term in the residual SAE to ensure it contributes only when meaningful domain-specific features are detected. This prevents unnecessary corrections to the pretrained SAE's output. 

We train the residual SAE on domain-specific data, taking the reconstruction error of the pretrained SAE as the training target. The loss function remains the standard SAE objective
\(
\mathcal{L} = \|\boldsymbol{e} - \hat{\boldsymbol{e}} \|_2^2 + \lambda \,\mathcal{L}_\text{reg},
\)
where the first term is the reconstruction loss, and the second term is the sparsity penalty scaled by $\lambda$. This setup allows the residual SAE to focus on features the pretrained SAE fails to capture, especially domain-specific ones. By using the same objective but a different target, it augments the pretrained SAE without competing with its existing features.

\subsection{Combined Model for Inference}
\label{sec:stitched_inference}

\begin{figure}
    \centering
    \includegraphics[width=0.7\linewidth]{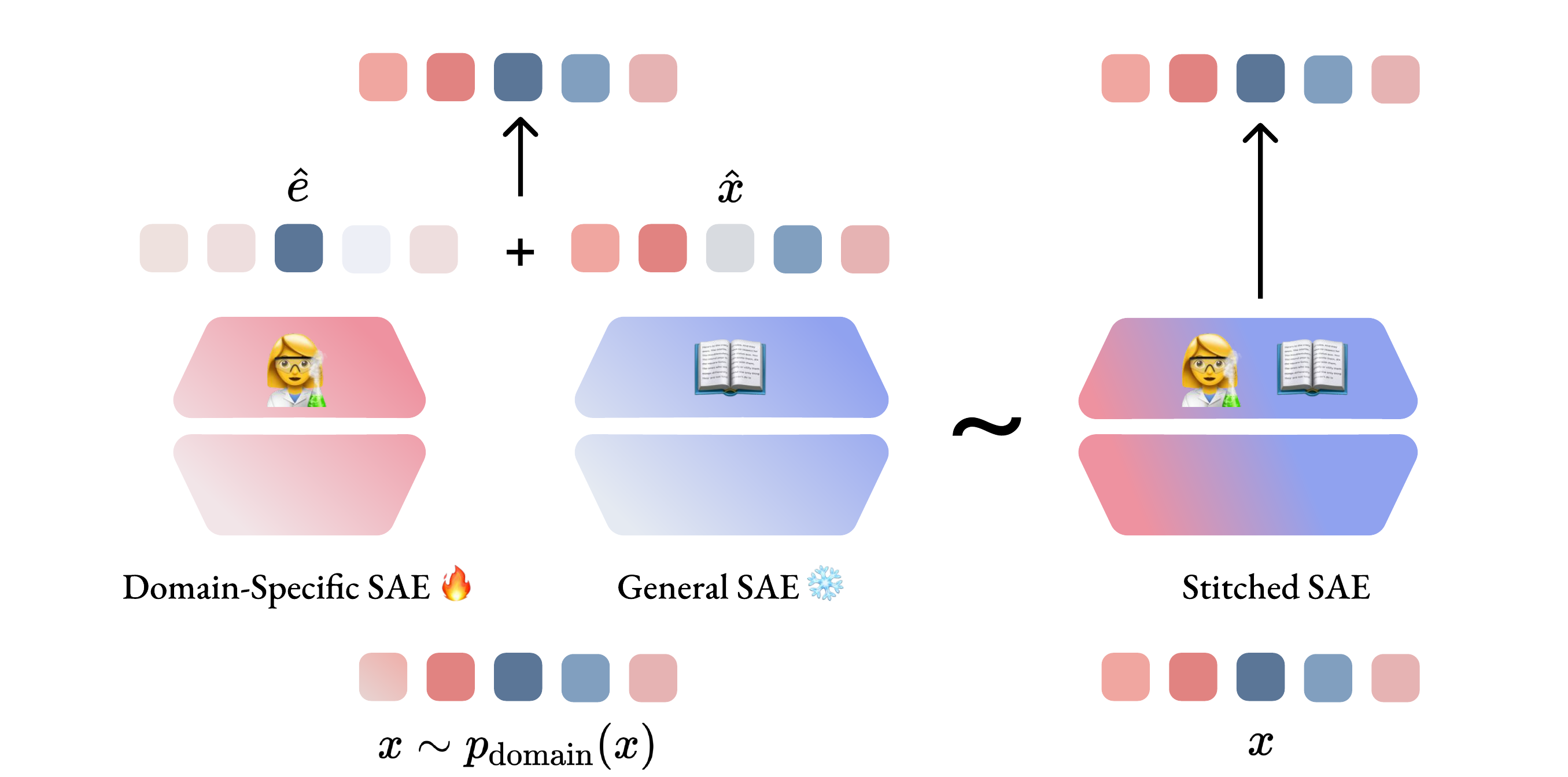}
    \caption{\textbf{Left}: During training, the domain-specific SAE is trained on the residuals of the general SAE (frozen weights). 
    \textbf{Right}: During inference, the domain-specific and general SAEs are stitched together. 
    We show the equivalence of the training (left) and inference (right) outputs in Section \ref{sec:stitched_inference}.}
    \label{fig:schematic}
\end{figure}

During inference, we combine outputs from the pretrained SAE and the domain-specific residual SAE as
\(
\vx = \hat{\vx} + \ve = \hat{\vx} + \hat{\ve} + \ve_2 \approx \hat{\vx} + \hat{\ve},
\)
where $\ve_2$ is the remaining error not captured by the residual SAE.

When multiple domain-specific residual SAEs are used, the combined error becomes
\(
\ve = \sum_{i=1}^{N} \hat{\ve}^{(i)} + \ve_N,
\)
where $N$ is the number of residual SAEs (each trained on a distinct domain) and $\hat{\ve}^{(i)}$ is the $i$-th residual SAE's reconstruction. The final reconstruction is then
\(
\vx  \approx \hat{\vx} + \sum_{i=1}^{N} \hat{\ve}^{(i)}.
\)

The resulting approach resembles tuning specific newly initialized features, although it only targets errors made by the pretrained SAE, thus providing complementary features. See Figure \ref{fig:schematic} for a schematic view of SAE Boost.

\section{Experiments}

We conducted extensive experiments to evaluate the effectiveness of our residual SAE approach across various domains and model architectures. Our experiments aim to answer the following questions: can residual SAEs effectively capture domain-specific features missed by general-purpose SAEs, does incorporating residual SAEs maintain performance on general domain texts, and does the approach generalize across different domains and base models?

\subsection{Experimental Setup}

\paragraph{Models.}
We conducted our experiments using two foundation models. We extracted the residual stream after the 24th transformer block of the Qwen-2.5-7B-Base \citep{qwen2.5} model and then trained our SAEs on these activations. The 24th layer was selected as a representative mid-to-late layer, where models typically develop rich feature representations. We also tested our approach using a publicly available pretrained SAE for the Llama-3.1-8B-Base model \citep{grattafiori2024llama3herdmodels} from the Llama Scope project \citep{he2024llamascope}, specifically the Llama3.1-8B-Base-L24R-8x, to verify generalizability.

\paragraph{Datasets.}
For general domain training, we used the Fineweb-edu corpus \citep{penedo2024finewebdatasetsdecantingweb}, a diverse collection of educational and informational web content. For domain-specific experiments, we selected three distinct domains: general texts in the \textbf{Russian language} from the Fineweb-2 dataset \citep{penedo2024finewebdatasetsdecantingweb}, representing a cross-lingual domain where certain features may be underrepresented in English-dominated pretraining; specialized scientific content focused on \textbf{chemistry data} \citep{li2023camel}, containing domain-specific terminology and concepts; and political discourse from \textbf{UN debates} \citep{hendersonkrass2022pileoflaw}, featuring formal diplomatic language and policy discussions.

\paragraph{Training Details.}
We trained our models using the following procedure. First, we trained an SAE on the Fineweb-edu corpus with standard SAE training objectives, including reconstruction loss and an L1 sparsity penalty. We then trained a Residual SAE on each domain-specific dataset to capture the reconstruction error of the pretrained SAE for that domain. For sparsity enforcement, we employed the batch topk activation \citep{bussmann2024batchtopksparseautoencoders} function with $k=50$ for the pretrained SAE and $k=5$ for the Residual SAE. During inference, we converted batch topk to jumpReLU by determining suitable activation thresholds on the respective training datasets.

\paragraph{Evaluation Metrics.}
We evaluated our models using three main metrics. \textbf{Explained Variance (EV)} measures the variance between the original activations and the SAE reconstructions, indicating the proportion of variance captured by the model. \textbf{LLM Cross-Entropy (LLM CE)} quantifies the change in cross-entropy when the original LLM activations are replaced with SAE reconstructions, thereby assessing how well the SAE preserves information relevant to the LLM’s next-token predictions. Finally, \textbf{L0 Sparsity} refers to the average number of active features per input, reflecting the model’s capacity utilization.

\begin{table}[t]
\centering
\small
\begin{tabular}{r|l|ccc}
\toprule
\textbf{Domain} & \textbf{Model} & \textbf{EV} $\uparrow$ & \textbf{LLM CE} $\downarrow$ & \textbf{L0 Sparsity} \\
\midrule

\multirow{4}{*}{Chemistry Data} 
& SAE (Qwen)          & 0.571 & 0.935 & 47 \\
& SAE Boost (Qwen)       & 0.716 (+25.39\%) & 0.767 (-17.97\%) & 52 \\
& SAE (Llama Scope)   & 0.551 & 1.208 & 52 \\
& SAE Boost (Llama Scope) & 0.702 (+27.40\%) & 0.995 (-17.63\%) & 57 \\
\midrule

\multirow{4}{*}{Russian Texts}
& SAE (Qwen)          & 0.455 & 4.716 & 57 \\
& SAE Boost (Qwen)       & 0.725 (+59.34\%) & 2.060 (-56.32\%) & 62 \\
& SAE (Llama Scope)   & 0.657 & 2.538 & 54 \\
& SAE Boost (Llama Scope) & 0.713 (+8.52\%) & 2.411 (-5.00\%) & 59 \\
\midrule

\multirow{4}{*}{UN Debates}
& SAE (Qwen)          & 0.693 & 2.464 & 56 \\
& SAE Boost (Qwen)       & 0.774 (+11.35\%) & 2.279 (-7.88\%) & 61 \\
& SAE (Llama Scope)   & 0.668 & 2.350 & 55 \\
& SAE Boost (Llama Scope) & 0.756 (+13.17\%) & 2.240 (-4.68\%) & 60 \\
\bottomrule
\end{tabular}
\caption{Performance of the pretrained SAE (baseline) and our SAE Boost approach across three specialized domains, each using a different pretrained LLM as the backbone. Our results show consistent, significant improvements in both reconstruction quality and LLM performance with SAE Boost. These findings confirm that our residual approach effectively captures domain-specific features that general-purpose SAEs miss, while maintaining a reasonable sparsity overhead. Furthermore, the proposed method generalizes well across various LLMs.}
\label{tab:domain_specific}
\end{table}

\paragraph{Baselines.}
We compared our residual SAE approach against several alternative methods for domain adaptation of SAEs, specifically: \textbf{Extended SAE (most active init)}, which extends the dictionary with new features initialized from the most active features on domain data; \textbf{Extended SAE (random init)}, which extends the dictionary using randomly initialized new features; \textbf{SAE Stitching}, which involves fully fine-tuning a pretrained SAE, identifying the features that changed the most (based on cosine similarity), and then stitching these features back into the original model; and \textbf{Full Fine-tuning}, for which SAE weights are simply fine-tuned using a specific domain dataset. For the Extended SAE approaches, only the newly added features were trained.

To ensure a fair comparison between our proposed SAE Boost approach and these baselines, we maintained consistent training conditions across all experiments. Each method was trained on the same number of tokens (1B for each domain), and each approach added the same number of features to the base model. Specifically, for the Extended SAE approaches, we added exactly the same number of features as our residual SAE. The SAE Boost approach added a residual dictionary of size 1024, while the Extended SAE approaches expanded the base SAE dictionary by the same number of features. The SAE Stitching approach selected 1024 features from the fine-tuned model for integration into the base model.

\subsection{Results}
\subsubsection{Domain-Specific Performance}

Table \ref{tab:domain_specific} presents the improvements in reconstruction quality and LLM cross-entropy on the domain-specific test sets when comparing our SAE Boost approach to the pretrained SAE alone. The results show consistent improvements across all three domains, with the residual SAE effectively capturing domain-specific features that the pretrained SAE missed. Moreover, the SAE Boost approach demonstrates these gains across different pretrained SAEs, indicating its generalizability across diverse base models.

\subsubsection{Impact on General Domain Performance}

A key concern when adapting models to specific domains is the potential degradation of general domain performance. Table \ref{tab:general_domain} shows the impact of incorporating the residual SAE on performance with general domain texts. Our results indicate that incorporating domain-specific residual SAEs has minimal impact on general domain performance, with changes of less than 1\% across all metrics. This confirms that our approach learns complementary features rather than competing with existing ones. As with previous results, we observed this behavior across different base models, which further supports the general capabilities of our method regardless of the underlying pretrained model. This consistency underscores its robustness and practical applicability.

\begin{table}[ht]
\centering
\small
\begin{tabular}{r|r|r|ccc}
\toprule
\textbf{SAE} & \textbf{Configuration} & \textbf{Domain} & \textbf{EV} $\uparrow$ & \textbf{LLM CE} $\downarrow$ & \textbf{L0 Sparsity} \\
\midrule
\multirow{4}{*}{\textbf{Qwen}} 
 & Baseline & -- 
 & 0.719 & 2.385 & 50 \\
 & \multirow{3}{*}{With Residual} 
 & Chemistry & 0.717 (-0.28\%) & 2.390 (+0.21\%) & 51 \\
 &  & Russian   & 0.719 (0.00\%) & 2.385 (0.00\%) & 50 \\
 &  & UN Debates & 0.719 (0.00\%) & 2.386 (+0.04\%) & 50 \\
\midrule
\multirow{4}{*}{\textbf{LLama}} 
 & Baseline & -- 
 & 0.672 & 2.599 & 55 \\
 & \multirow{3}{*}{With Residual} 
 & Chemistry & 0.672 (+0.00\%) & 2.596 (-0.12\%) & 55 \\
 &  & Russian   & 0.676 (+0.60\%) & 2.612 (+0.50\%) & 55 \\
 &  & UN Debates & 0.671 (-0.15\%) & 2.585 (-0.54\%) & 56 \\
\bottomrule
\end{tabular}
\caption{The results of adding domain-specific SAE Boost on performance in general domain tasks. We compare the baseline (without SAE Boost) to configurations that use SAE Boost from three different domains. The results show negligible impact on general domain performance across all metrics. This finding confirms that our approach effectively isolates domain-specific features without compromising general capabilities, demonstrating that residual SAEs learn complementary rather than competing features. The evaluation was conducted using both Qwen SAEs and LLammaScope.}
\label{tab:general_domain}
\end{table}

\begin{table}[ht]
\centering
\small
\begin{tabular}{l|cc|cc}
\toprule
\textbf{Method} & \textbf{General EV}  $\uparrow$& \textbf{General L0} & \textbf{UN Debates EV} $\uparrow$ & \textbf{UN Debates L0} \\
\midrule
SAE Boost (ours) & 0.719 & 50 & 0.774 & 61 \\
Extended SAE (most act) & 0.714& 52 & 0.770 & 64 \\
Extended SAE (random) & 0.716 & 52 & 0.774 & 64 \\
SAE Stitching & 0.719 & 50 & 0.703 & 57 \\
Full fine-tuning & 0.515 & 81 & 0.850 & 50 \\

\bottomrule
\end{tabular}
\caption{Comparison of our SAE Boost approach against alternative methods for domain adaptation. While all methods maintain similar general domain performance, SAE Boost achieves the best domain-specific performance with efficient sparsity. Extended SAE approaches demonstrate competitive domain performance but require higher sparsity, whereas SAE Stitching significantly underperforms on domain adaptation. Full fine-tuning suffers from catastrophic forgetting. These results illustrate that our approach strikes the optimal balance between preserving general capabilities, enhancing domain performance, and efficient feature utilization. Additional detailed comparisons for the Chemistry and Russian domains can be found in Tables~\ref{tab:chemistry-adaptation} and \ref{tab:russian-adaptation} in the Appendix, confirming consistent patterns across diverse domains.}
\label{tab:adaptation-methods}
\end{table}

\begin{table}[ht]
\small
\centering
\begin{tabular}{l|cc|cc}
\toprule
\textbf{Method} & \textbf{General EV} $\uparrow$ & \textbf{General L0} & \textbf{UN Debates EV} $\uparrow$ & \textbf{UN Debates L0} \\
\midrule
Baseline (No adaptation) & 0.719 & 50 & 0.695 & 56 \\
\midrule
\multicolumn{5}{l}{\textit{Multi-Domain Approaches:}} \\
\midrule
SAE Boost & \textbf{0.715} & 52 & \textbf{0.770} & 62 \\
Extended SAE (random) & 0.708 & 55 & 0.760 & 67 \\
SAE Stitching & 0.674 & 53 & 0.670 & 59 \\
\bottomrule
\end{tabular}
\caption{Performance on the UN Debates domain with single vs.\ multiple domain adaptations applied simultaneously. This table demonstrates that even when applying all three domain-specific models together (Chemistry, Russian, and UN Debates), our SAE Boost approach maintains excellent domain-specific performance on UN Debates compared to single-domain adaptation, with minimal impact on general domain performance. In contrast, alternative approaches show more significant trade-offs in either sparsity, general performance, or domain performance.}
\label{tab:multi_domain_comparison}
\end{table}

\subsubsection{Comparison of Domain Adaptation Methods}
\label{sec:other_methods}

As shown in Table~\ref{tab:adaptation-methods}, our SAE Boost approach achieves the best balance between domain-specific performance and overall general capabilities among the compared methods for domain adaptation of SAEs. While the Extended SAE approaches exhibit competitive performance on the UN Debates domain, they require slightly higher L0 sparsity, indicating less efficient feature utilization. SAE Stitching performs comparably to SAE Boost on the general domain but demonstrates notably weaker domain-specific adaptation. Full fine-tuning showed a large Explainable Variance score, yet it largely forgets information with a general domain dataset.

We hypothesize that Extended SAE loses performance in the general domain due to the concurrency of newly trained features with old ones (i.e., we cannot ensure that they precisely cover the new domain). In contrast, SAE Stitching fails to capture newly emerging features from domain-specific datasets, possibly because most changed features are initialized with features previously trained to capture general-domain data. Both approaches demonstrate misalignment between new and old features, aligning with the motivation described in Section~\ref{sec:intro}. Consequently, our approach offers a key advantage: it maintains strong general-domain performance while achieving superior domain-specific adaptation at efficient sparsity levels. Similar patterns emerge in other domains, with additional results for Chemistry data and Russian texts presented in Tables~\ref{tab:chemistry-adaptation} and \ref{tab:russian-adaptation} in the Appendix. These evaluations further confirm the effectiveness and generalizability of our approach across diverse domain types.

\begin{figure}[ht]

    \centering
    \includegraphics[width=0.8\linewidth]{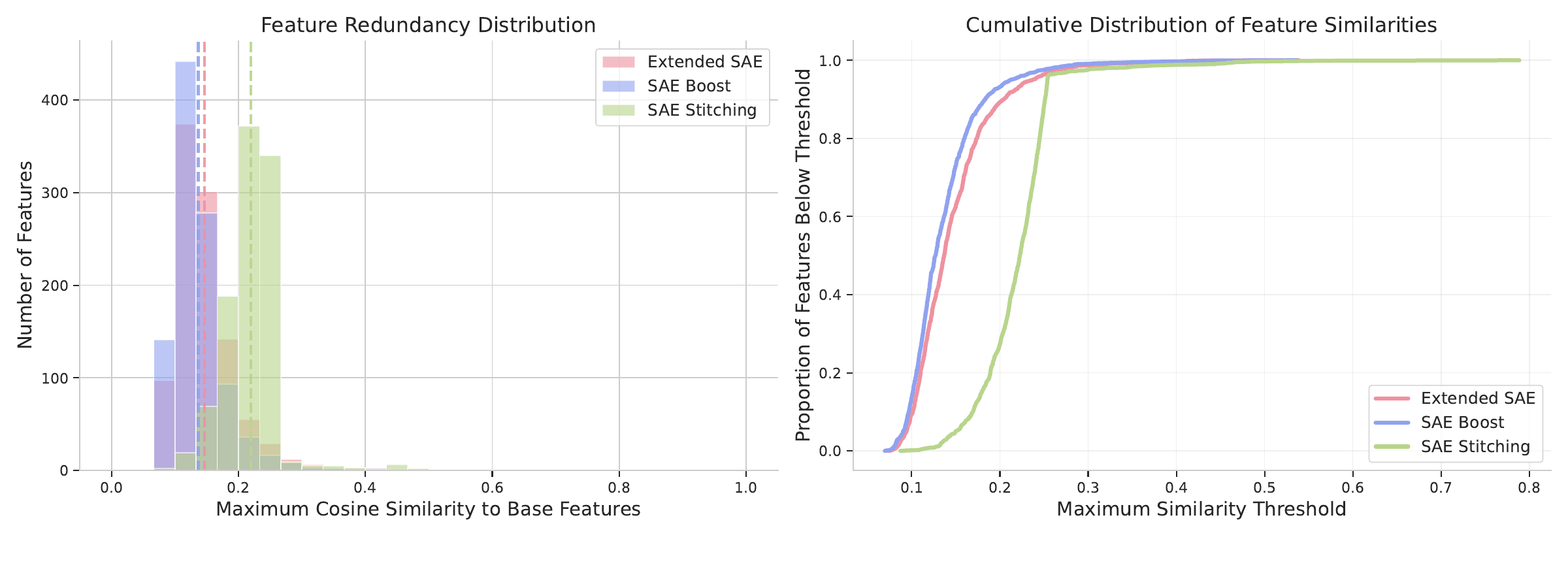}
\caption{
Similarity analysis between domain-specific UN Debate features and the base features. (left) The distribution of the maximum cosine similarity between each domain-adapted feature and any feature in the base model. (right) The cumulative distribution function of these similarities. Notably, SAE Boost features exhibit lower similarity to base features, indicating that they capture more novel domain-specific information rather than merely replicating existing representations.
}
    \label{fig:debates-sim}
\end{figure}

To understand performance differences, we analyze feature similarity between domain-adapted and pretrained features. Figure \ref{fig:debates-sim} shows that SAE Boost features exhibit lower cosine similarity to base features than both Extended SAE and SAE Stitching. These similarity patterns explain our method's advantages: Extended SAE features overlap more with base features, potentially causing interference during inference, while SAE Stitching struggles to capture novel concepts due to its initialization. In contrast, by explicitly targeting reconstruction errors, SAE Boost learns more complementary features that effectively address representational gaps in the base model.

\begin{figure}[hbtp]
    \centering
    \includegraphics[width=0.5\linewidth]{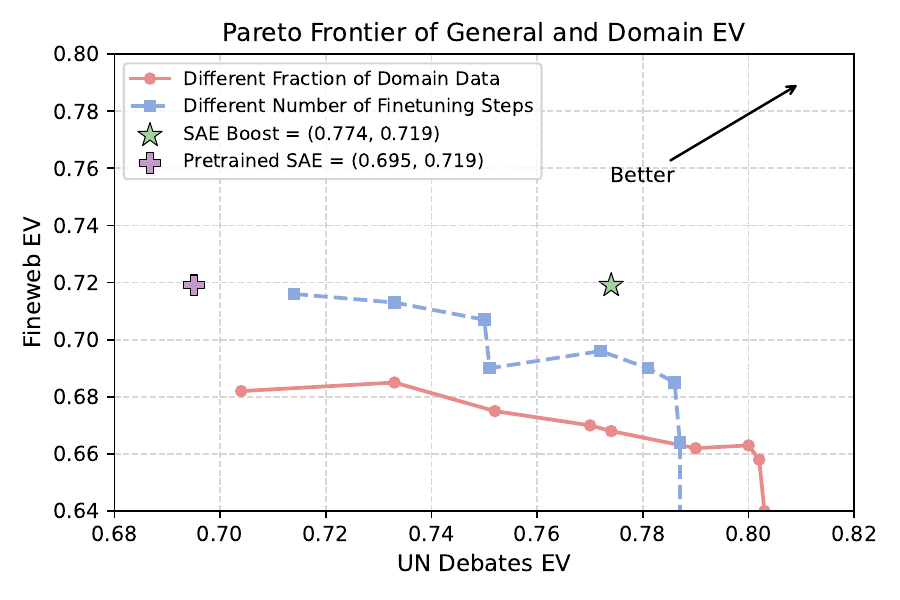}
    \caption{SAE Boost is compared to other baselines on the Pareto frontier, with the optimal area marked by an arrow. SAE Boost is closer to the optimal point. See Section \ref{sec:other_methods} for more details.} 
    \label{fig:pareto}
\end{figure}

We also compare SAE Boost to simple baselines that involve fine-tuning with varying numbers of training steps and different data mixtures. Note that SAE Boost does not require general-domain data. We measure explained variance on general-domain texts from the FineWeb dataset and domain-specific texts from the UN Debates dataset. The results are presented in Figure \ref{fig:pareto}. The proposed method is closer to the optimal point on the Pareto frontier.

\subsubsection{Multi-Domain Adaptation}

A significant advantage of our residual approach is its modularity and the ability to incorporate multiple domain-specific adaptations simultaneously without compromising performance. To demonstrate this, we trained separate SAE Boosts for the Chemistry, Russian, and UN Debates domains, then evaluated their combined performance when applied concurrently.

Table~\ref{tab:multi_domain_comparison} presents the results obtained when evaluating on the UN Debates domain, while Tables~\ref{tab:multi-chemistry} and~\ref{tab:multi-russian} in the Appendix show corresponding results for the Chemistry and Russian domains, respectively. These results illustrate that even when all three domain-specific residual SAEs are applied simultaneously, our approach maintains excellent performance compared to both single-domain adaptation and alternative methods.

\subsubsection{Training Dynamics and Feature Convergence}

\begin{figure}[ht!]
    \centering
    \includegraphics[width=0.6\linewidth]{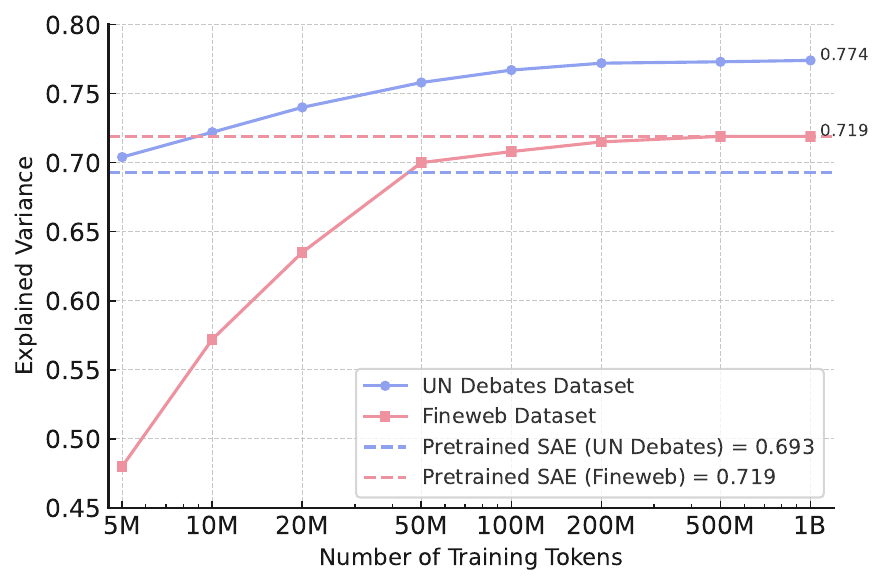}
\caption{Evolution of explained variance during training for domain-specific and general domains. This figure tracks the training progress of the residual SAE on both UN Debates (blue line) and Fineweb datasets (orange line), measured against the baseline performance of the pretrained SAE (dashed lines). This demonstrates that sufficient training ($>$ 50M tokens) is critical for the residual SAE to develop complementary features that enhance domain-specific performance without degrading general capabilities. The training dynamics reveal that undertraining the residual SAE ($<$50M tokens) would result in suboptimal feature quality with potential negative impacts on general domain performance.}
    \label{fig:ev-training}
\end{figure}

Figure~\ref{fig:ev-training} illustrates the explained variance (EV) as a function of training tokens for our residual SAE on domain-specific data. Undertraining the residual SAE produces poorly defined features that degrade the pretrained SAE’s performance on general domain tasks. We found that residual SAEs trained on fewer than 100M tokens can reduce general domain performance by up to 31\% in explained variance. As training progresses beyond the 200M token mark, the residual features become more distinct and complementary to those in the pretrained SAE. At this point, we observe minimal interference with general domain performance (less than 1\% change in explained variance), as shown in Table~\ref{tab:general_domain}. This finding suggests that well-converged features from the residual SAE are crucial for maintaining general domain capabilities while enhancing domain-specific performance.

These observations have important implications for practical deployment of our approach: (1) residual SAEs must be sufficiently trained to avoid degrading general performance, (2) practitioners should monitor both domain-specific improvements and general domain performance during training, and (3) early stopping based solely on domain-specific metrics may lead to suboptimal feature quality.

\subsubsection{Interpretability Analysis}

To demonstrate the effectiveness of our SAE Boost approach in capturing domain-specific features, we performed a detailed interpretability analysis of the features learned by both the pretrained SAE and the domain-specific SAE Boost models. This analysis reveals how SAE Boost successfully identifies meaningful domain-specific concepts that general-purpose SAEs might miss.

\begin{figure}[t]
    \centering
    \includegraphics[width=0.8\linewidth]{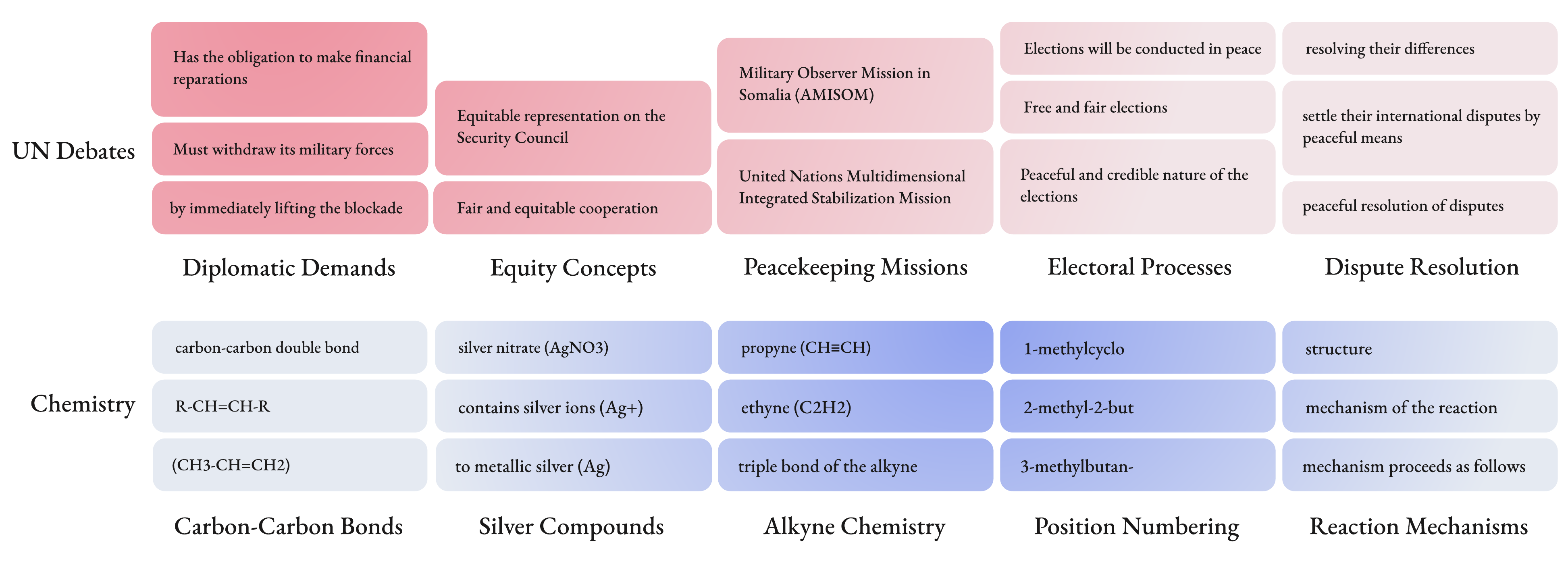}
\caption{Representative domain-specific features discovered by SAE Boost, along with their corresponding top activations.}
    \label{fig:features}
\end{figure}

We analyzed the learned features by examining their top activations across domain-specific corpora. For each feature, we collected text segments that produced the highest activation values, allowing us to interpret the semantic concepts captured by that feature. Figure~\ref{fig:features} presents representative domain-specific features discovered by our SAE Boost models in different domains, illustrating the distinct concepts captured in each. This table demonstrates how our SAE Boost approach effectively identifies domain-specific concepts across diverse domains. In the UN Debates domain, features capture diplomatic terminology, policy frameworks, and peacekeeping operations. Chemistry features identify specific chemical compounds, reaction types, and nomenclature conventions. Together, these features show that our approach can discover meaningful domain-specific concepts that might be overlooked by general-purpose SAEs.

\begin{figure}[t]
    \centering
    \includegraphics[width=0.8\linewidth]{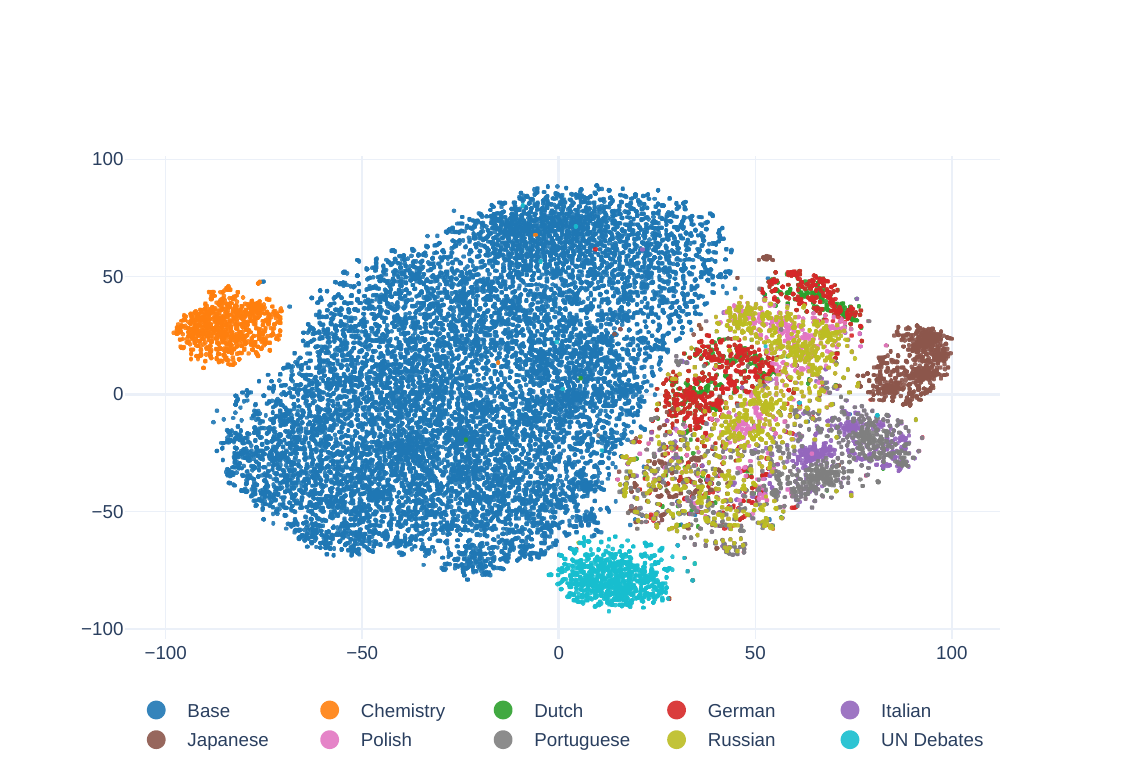}
\caption{A t-SNE visualization of the feature embeddings from the base SAE and multiple domain-specific SAE Boost models. The figure reveals distinct clustering by domain type, with base features (blue) occupying the central region, Chemistry features (orange) forming a compact cluster to the left, and UN Debates features (turquoise) at the bottom. Notably, cross-lingual features cluster together on the right side, with related languages in close proximity (e.g., the Romance languages Portuguese and Italian, and the Germanic languages German and Dutch). This organization demonstrates that SAE Boost captures both domain-specific knowledge and inherent relationships between related domains, while maintaining clear separation from general features.}
    \label{fig:tsne_multilingual}
\end{figure}

To better understand the global organization of features across domains, we conducted t-SNE visualizations of the feature embeddings from our SAE Boost models. These visualizations offer insights into how domain-specific features relate to both general features and each other. Figure~\ref{fig:tsne_multilingual} presents a t-SNE visualization of feature embeddings from the base SAE and multiple domain-specific SAE Boost models. The figure reveals distinct clustering by domain type, with base features (blue) occupying the largest central region while domain-specific features form distinct clusters at the periphery. This arrangement demonstrates that our SAE Boost approach identifies features that are not only semantically distinct (as shown by our activation analysis) but also structurally distinct in the embedding space. Particularly noteworthy is the organization of cross-lingual features on the right side of the visualization. We observe that features from related language families tend to cluster together: the Slavic languages (Russian and Polish) form adjacent clusters, as do the Romance languages (Italian and Portuguese) and the Germanic languages (German and Dutch). Japanese, being linguistically unrelated to the European languages, forms its own distinct cluster. This organization reflects linguistic reality, suggesting that our SAE Boost approach naturally captures underlying typological relationships between languages. The clear separation between domains in these visualizations provides further evidence that our SAE Boost approach effectively captures domain-specific knowledge in distinct, interpretable feature spaces.

\section{Conclusion}

This paper introduces SAE Boost, a residual learning approach that addresses feature blindness in the sparse autoencoders used for LLM interpretability. By training a secondary SAE to specifically model the reconstruction error of a pretrained SAE on domain-specific texts, SAE Boost enables more comprehensive interpretability across specialized domains without requiring full retraining.

Our experiments show that SAE Boost improves both reconstruction quality and LLM cross-entropy across multiple domains, while preserving overall performance. The key advantage of this method is selective domain enhancement without modifying the base model.

SAE Boost represents an important step forward in the mechanistic interpretability of LLMs by allowing researchers to selectively enhance their tools for particular domains of interest. As language models continue to grow in capability and application scope, such targeted approaches will become increasingly valuable for understanding their internal representations and behaviors.

\bibliography{colm2025_conference}
\bibliographystyle{colm2025_conference}

\appendix
\section{Additional Experimental Results}

\subsection{Cross-Lingual Domain Performance}
Table~\ref{tab:cross-lingual} presents additional cross-lingual domain performance results, which demonstrate the effectiveness of our residual SAE approach across various languages.
\begin{table}[ht]
\centering

\begin{tabular}{r|cc|ccc}
\toprule
\textbf{Language} & Pretrain EV $\uparrow$ & Pretrain L0 & Residual EV $\uparrow$ & Residual L0 \\
\midrule
Russian     & 0.455 & 57 & 0.725 & 62 \\
Japanese    & 0.348 & 58 & 0.695 & 63 \\
Portuguese  & 0.532 & 54 & 0.735 & 59 \\
Italian     & 0.543 & 53 & 0.742 & 58 \\
Dutch       & 0.568 & 48 & 0.744 & 53 \\
Polish      & 0.510 & 52 & 0.748 & 57 \\
German      & 0.559 & 52 & 0.733 & 57 \\
\bottomrule
\end{tabular}
\caption{Cross-lingual performance of the residual SAE approach.}
\label{tab:cross-lingual}
\end{table}

These findings further support our primary conclusion that the residual SAE effectively captures domain-specific features across multiple languages. Notably, Japanese exhibits the greatest improvement, likely due to its distinct writing system and linguistic characteristics compared to the predominantly English-based training corpus.

\subsection{Comparison of Domain Adaptation Methods}

Tables~\ref{tab:chemistry-adaptation} and \ref{tab:russian-adaptation} provide detailed comparisons between our SAE Boost approach and alternative methods for domain adaptation on the chemistry and Russian domains, respectively.

\begin{table}[ht]
\centering
\begin{tabular}{r|cc|cc}
\toprule
\textbf{Method} & General EV $\uparrow$ & General L0 & Chemistry EV $\uparrow$ & Chemistry L0 \\
\midrule
SAE Boost (ours) & 0.719 & 50 & 0.716 & 52 \\
Extended SAE & 0.714 & 52 & 0.726 & 56 \\
SAE Stitching & 0.715 & 50 & 0.590 & 49 \\
Full fine-tuning & 0.448 & 95 & 0.837 & 50  \\
\bottomrule
\end{tabular}
\caption{Comparison of domain adaptation methods on chemistry data.}
\label{tab:chemistry-adaptation}
\end{table}

Although the Extended SAE approach achieves slightly better performance on chemistry data, it requires higher sparsity and slightly reduces general domain performance. In contrast, our SAE Boost approach strikes a better balance between domain-specific and general domain capabilities.

\begin{table}[ht]
\centering
\begin{tabular}{r|cc|cc}
\toprule
\textbf{Method} & General EV $\uparrow$ & General L0 & Russian EV $\uparrow$ & Russian L0 \\
\midrule
SAE Boost (ours) & 0.719 & 50 & 0.725 & 62 \\
Extended SAE & 0.708 & 52 & 0.741 & 71 \\
SAE Stitching & 0.702 & 50 & 0.505 & 60 \\
Full fine-tuning & 0.517 & 66 & 0.829 & 50  \\
\bottomrule
\end{tabular}
\caption{Comparison of domain adaptation methods on Russian texts.}
\label{tab:russian-adaptation}
\end{table}

For Russian texts, although the Extended SAE achieves slightly better domain-specific performance, it requires significantly higher sparsity (71 vs. 62), suggesting less efficient feature usage. Meanwhile, SAE Boost retains stronger general domain performance while achieving competitive domain-specific results.

\subsection{Multi-Domain Adaptation}

Tables~\ref{tab:multi-chemistry} and \ref{tab:multi-russian} show the results of simultaneously applying multiple domain-specific residual SAEs to the chemistry and Russian domains, respectively. These experiments demonstrate that our SAE Boost approach maintains robust performance even when multiple domain-specific residual SAEs are utilized concurrently.

\begin{table}[ht]
\centering

\begin{tabular}{r|cc|cc}
\toprule
\textbf{Method} & General EV $\uparrow$ & General L0 & Domain EV $\uparrow$ & Domain L0 \\
\midrule
\multicolumn{5}{l}{Single-Domain Approaches:} \\
Baseline (No adaptation) & 0.719 & 50 & 0.571 & 47 \\
SAE Boost (Chemistry only) & 0.719 & 50 & 0.716 & 52 \\
\midrule
\multicolumn{5}{l}{Multi-Domain Approaches:} \\
SAE Boost (3 domains) & 0.712 & 52 & 0.711 & 54 \\
Extended SAE (3 domains) & 0.704 & 55 & 0.712 & 61 \\
SAE Stitching (3 domains) & 0.685 & 53 & 0.510 & 51 \\
\bottomrule
\end{tabular}
\caption{Multiple domain adaptation results on chemistry data.}
\label{tab:multi-chemistry}
\end{table}

In the chemistry domain, our multi-domain SAE Boost approach (with three domains) achieves performance similar to single-domain adaptation (0.711 vs.\ 0.716 EV) with only a small increase in sparsity. Although the Extended SAE method obtains comparable domain performance, it incurs higher sparsity overhead, while SAE Stitching experiences considerable performance degradation under multi-domain conditions.

\begin{table}[ht]
\centering

\begin{tabular}{r|cc|cc}
\toprule
\textbf{Method} & General EV $\uparrow$ & General L0 & Domain EV $\uparrow$ & Domain L0 \\
\midrule
\multicolumn{5}{l}{Single-Domain Approaches:} \\
Baseline (No adaptation) & 0.720 & 50 & 0.455 & 57 \\
SAE Boost (Russian only) & 0.719 & 50 & 0.725 & 62 \\
\midrule
\multicolumn{5}{l}{Multi-Domain Approaches:} \\
SAE Boost (3 domains) & 0.713 & 52 & 0.714 & 66 \\
Extended SAE (3 domains) & 0.701 & 55 & 0.720 & 81 \\
SAE Stitching (3 domains) & 0.677 & 53 & 0.503 & 62 \\
\bottomrule
\end{tabular}
\caption{Multiple domain adaptation results on Russian texts.}
\label{tab:multi-russian}
\end{table}

For Russian texts, our multi-domain SAE Boost approach preserves strong performance (0.714 vs.\ 0.725 EV) with only a moderate increase in sparsity. Although the Extended SAE method exhibits competitive domain performance, it demands substantially higher sparsity (81 vs.\ 66). In contrast, SAE Stitching again underperforms in the multi-domain scenario.

Overall, these findings underscore the primary advantage of our approach: it preserves high performance in the general domain while delivering superior domain-specific adaptation with efficient sparsity across multiple domains.

\subsection{Interpretability on Domain-Specific Data}

To support our interpretability claims. While Figure 4 provides anecdotal support, we further substantiate interpretability using quantitative metrics following \citet{paulo2024automatically}. We evaluated all 1024 residual SAE features on the chemistry dataset and randomly sampled $1024$ base SAE features evaluated on the FineWeb-Edu dataset. Results are presented in Table \ref{tab:sae_scores}. SAE Boost consistently outperforms base SAE which indicates presence of interpretable domain-specific features.

\begin{table}[ht]
\centering
\begin{tabular}{r|cc}
\toprule
\textbf{SAE} & \textbf{Mean (median) detection score} & \textbf{Mean (median) fuzzing score} \\
\midrule
Base SAE & 0.67 (0.65) ± 0.16 & 0.64 (0.6) ± 0.12 \\
Residual SAE & 0.75 (0.77) ± 0.17 & 0.68 (0.68) ± 0.12 \\
\bottomrule
\end{tabular}
\caption{Detection and fuzzing scores for different SAE variants}
\label{tab:sae_scores}
\end{table}

\subsection{Domain-Specific Features Analysis}
\label{ap:domain-specific}

To ensure that SAE Boost discovers domain-specific features not represented in the base SAE, we selected five features from our method and compared them to the top three most similar features (by cosine similarity) in the base SAE on the chemistry dataset. The results are presented in Table \ref{tab:sae_boost_mapping}. SAE Boost indeed discovers new features related to chemistry.

\begin{table}[ht]
\centering
\begin{tabular}{r|l}
\toprule
\textbf{SAE Boost Feature} & \textbf{Most Similar General‑Domain Features} \\
\midrule
\multirow{3}{*}{\shortstack[l]{Oxygen in chemical contexts}} &
(1) Numerical thresholds;\\
& (2) Names of people;\\
& (3) Risk‑related phrases \\
\midrule
\multirow{3}{*}{\shortstack[l]{Chemical and thermal stability}} &
(1) Durability references;\\
& (2) Legal protections;\\
& (3) Safety responsibilities \\
\midrule
\multirow{3}{*}{\shortstack[l]{Supramolecular structure formation}} &
(1) Protein databases;\\
& (2) Scientific measurements;\\
& (3) Mineralogy terms \\
\midrule
\multirow{3}{*}{\shortstack[l]{Chemistry energy values}} &
(1) Chemical reaction terms;\\
& (2) Calibration concepts;\\
& (3) Physics discussions \\
\midrule
\multirow{3}{*}{\shortstack[l]{Hydrogen chemistry}} &
(1) Personal relationships;\\
& (2) Compound applications;\\
& (3) Synthetic processes \\
\bottomrule
\end{tabular}
\caption{Mapping of SAE Boost Features to similar General‑Domain Features. See Appendix \ref{ap:domain-specific} for more details.}
\label{tab:sae_boost_mapping}
\end{table}

\subsection{SAE Boost under different sparsity levels}

\begin{table}[htbp]
\centering
\begin{tabular}{r|cc|cc}
\toprule
\textbf{Top-$k$} & \textbf{General EV} & \textbf{General L0} & \textbf{Domain EV} & \textbf{Domain L0} \\
\midrule
5  & 0.719 & 50  & 0.774 & 61  \\
10 & 0.719 & 52  & 0.781 & 66  \\
20 & 0.719 & 56  & 0.788 & 76  \\
50 & 0.721 & 72  & 0.798 & 106 \\
\bottomrule
\end{tabular}
\caption{Performance metrics at different top‑$k$ values.}
\label{tab:topk_metrics}
\end{table}

To ensure that the choice of the top-$k$ value for the residual SAE is optimal, we conducted a sensitivity analysis, which indicated minor domain performance gains with higher $k$, but at the cost of reduced sparsity and interpretability. Results are presented in Table \ref{tab:topk_metrics}. We selected $k=5$ to balance strong domain performance, minimal general-domain disruption, and optimal interpretability.

\end{document}

%% file: math_commands.tex
\usepackage{amsmath,amsfonts,bm}

\def\eqref#1{equation~\ref{#1}}

\def\1{\bm{1}}

\def\vb{{\bm{b}}}

\def\ve{{\bm{e}}}

\def\vx{{\bm{x}}}

\def\mW{{\bm{W}}}

\DeclareMathAlphabet{\mathsfit}{\encodingdefault}{\sfdefault}{m}{sl}
\SetMathAlphabet{\mathsfit}{bold}{\encodingdefault}{\sfdefault}{bx}{n}













%% file: colm2025_conference.bbl
\begin{thebibliography}{21}
\providecommand{\natexlab}[1]{#1}
\providecommand{\url}[1]{\texttt{#1}}
\expandafter\ifx\csname urlstyle\endcsname\relax
  \providecommand{\doi}[1]{doi: #1}\else
  \providecommand{\doi}{doi: \begingroup \urlstyle{rm}\Url}\fi

\bibitem[Bengio(2013)]{bengio2013deep}
Yoshua Bengio.
\newblock Deep learning of representations: Looking forward.
\newblock In \emph{International Conference on Statistical Language and Speech Processing}, pp.\  1--37. Springer, 2013.

\bibitem[Bricken et~al.(2023)Bricken, Templeton, Batson, Chen, Jermyn, Conerly, Turner, Anil, Denison, Askell, et~al.]{bricken2023towards}
Trenton Bricken, Adly Templeton, Joshua Batson, Brian Chen, Adam Jermyn, Tom Conerly, Nick Turner, Cem Anil, Carson Denison, Amanda Askell, et~al.
\newblock Towards monosemanticity: Decomposing language models with dictionary learning.
\newblock \emph{Transformer Circuits Thread}, 2, 2023.

\bibitem[Bussmann et~al.(2024)Bussmann, Leask, and Nanda]{bussmann2024batchtopksparseautoencoders}
Bart Bussmann, Patrick Leask, and Neel Nanda.
\newblock Batchtopk sparse autoencoders, 2024.
\newblock URL \url{https://arxiv.org/abs/2412.06410}.

\bibitem[Cunningham et~al.(2023)Cunningham, Ewart, Riggs, Huben, and Sharkey]{cunningham2023sparse}
Hoagy Cunningham, Aidan Ewart, Logan Riggs, Robert Huben, and Lee Sharkey.
\newblock Sparse autoencoders find highly interpretable features in language models.
\newblock \emph{arXiv preprint arXiv:2309.08600}, 2023.

\bibitem[Gao et~al.(2024)Gao, Dupr{\'e}~la Tour, Tillman, Goh, Troll, Radford, Sutskever, Leike, and Wu]{gao2024scaling}
Leo Gao, Tom Dupr{\'e}~la Tour, Henk Tillman, Gabriel Goh, Rajan Troll, Alec Radford, Ilya Sutskever, Jan Leike, and Jeffrey Wu.
\newblock Scaling and evaluating sparse autoencoders.
\newblock \emph{arXiv preprint arXiv:2406.04093}, 2024.

\bibitem[He et~al.(2024)He, Shu, Ge, Chen, Wang, Zhou, Liu, Guo, Huang, Wu, et~al.]{he2024llamascope}
Zhengfu He, Wentao Shu, Xuyang Ge, Lingjie Chen, Junxuan Wang, Yunhua Zhou, Frances Liu, Qipeng Guo, Xuanjing Huang, Zuxuan Wu, et~al.
\newblock Llama scope: Extracting millions of features from llama-3.1-8b with sparse autoencoders.
\newblock \emph{arXiv preprint arXiv:2410.20526}, 2024.

\bibitem[Henderson* et~al.(2022)Henderson*, Krass*, Zheng, Guha, Manning, Jurafsky, and Ho]{hendersonkrass2022pileoflaw}
Peter Henderson*, Mark~S. Krass*, Lucia Zheng, Neel Guha, Christopher~D. Manning, Dan Jurafsky, and Daniel~E. Ho.
\newblock Pile of law: Learning responsible data filtering from the law and a 256gb open-source legal dataset, 2022.
\newblock URL \url{https://arxiv.org/abs/2207.00220}.

\bibitem[Higgins et~al.(2017)Higgins, Matthey, Pal, Burgess, Glorot, Botvinick, Mohamed, and Lerchner]{higgins2017beta}
Irina Higgins, Loic Matthey, Arka Pal, Christopher Burgess, Xavier Glorot, Matthew Botvinick, Shakir Mohamed, and Alexander Lerchner.
\newblock beta-{VAE}: Learning basic visual concepts with a constrained variational framework.
\newblock In \emph{International Conference on Learning Representations}, 2017.

\bibitem[Leask et~al.(2025)Leask, Bussmann, Pearce, Bloom, Tigges, Moubayed, Sharkey, and Nanda]{leask2025sparseautoencoderscanonicalunits}
Patrick Leask, Bart Bussmann, Michael Pearce, Joseph Bloom, Curt Tigges, Noura~Al Moubayed, Lee Sharkey, and Neel Nanda.
\newblock Sparse autoencoders do not find canonical units of analysis, 2025.
\newblock URL \url{https://arxiv.org/abs/2502.04878}.

\bibitem[Li et~al.(2023)Li, Hammoud, Itani, Khizbullin, and Ghanem]{li2023camel}
Guohao Li, Hasan Abed Al~Kader Hammoud, Hani Itani, Dmitrii Khizbullin, and Bernard Ghanem.
\newblock Camel: Communicative agents for "mind" exploration of large scale language model society, 2023.

\bibitem[Mallat \& Zhang(1993)Mallat and Zhang]{mallat1993matching}
St{\'e}phane~G Mallat and Zhifeng Zhang.
\newblock Matching pursuits with time-frequency dictionaries.
\newblock In \emph{Proceedings of IEEE International Conference on Acoustics, Speech and Signal Processing}, volume~3, pp.\  3397--3400. IEEE, 1993.

\bibitem[Muhamed et~al.(2024)Muhamed, Diab, and Smith]{muhamed2024decodingdarkmatterspecialized}
Aashiq Muhamed, Mona Diab, and Virginia Smith.
\newblock Decoding dark matter: Specialized sparse autoencoders for interpreting rare concepts in foundation models, 2024.
\newblock URL \url{https://arxiv.org/abs/2411.00743}.

\bibitem[Olah et~al.(2017)Olah, Mordvintsev, and Schubert]{olah2017feature}
Chris Olah, Alexander Mordvintsev, and Ludwig Schubert.
\newblock Feature visualization.
\newblock \emph{Distill}, 2\penalty0 (11):\penalty0 e7, 2017.

\bibitem[Olshausen \& Field(1996)Olshausen and Field]{olshausen1996emergence}
Bruno~A Olshausen and David~J Field.
\newblock Emergence of simple-cell receptive field properties by learning a sparse code for natural images.
\newblock \emph{Nature}, 381\penalty0 (6583):\penalty0 607--609, 1996.

\bibitem[Olshausen \& Field(1997)Olshausen and Field]{olshausen1997sparse}
Bruno~A Olshausen and David~J Field.
\newblock Sparse coding with an overcomplete basis set: A strategy employed by {V1}?
\newblock \emph{Vision Research}, 37\penalty0 (23):\penalty0 3311--3325, 1997.

\bibitem[Paulo et~al.(2024)Paulo, Mallen, Juang, and Belrose]{paulo2024automatically}
Gonçalo Paulo, Alex Mallen, Caden Juang, and Nora Belrose.
\newblock Automatically interpreting millions of features in large language models.
\newblock \emph{arXiv preprint arXiv: 2410.13928}, 2024.

\bibitem[Penedo et~al.(2024)Penedo, Kydlíček, allal, Lozhkov, Mitchell, Raffel, Werra, and Wolf]{penedo2024finewebdatasetsdecantingweb}
Guilherme Penedo, Hynek Kydlíček, Loubna~Ben allal, Anton Lozhkov, Margaret Mitchell, Colin Raffel, Leandro~Von Werra, and Thomas Wolf.
\newblock The fineweb datasets: Decanting the web for the finest text data at scale, 2024.
\newblock URL \url{https://arxiv.org/abs/2406.17557}.

\bibitem[Smith et~al.(2025)Smith, Rajamanoharan, Conmy, McDougall, Lieberum, Kramár, Shah, and Nanda]{saenanda}
Lewis Smith, Senthooran Rajamanoharan, Arthur Conmy, Callum McDougall, Tom Lieberum, János Kramár, Rohin Shah, and Neel Nanda.
\newblock Negative results for saes on downstream tasks and deprioritising sae research (gdm mech interp team progress update 2), 2025.
\newblock URL \url{https://www.alignmentforum.org/posts/4uXCAJNuPKtKBsi28/}.

\bibitem[team(2024)]{grattafiori2024llama3herdmodels}
LLaMA team.
\newblock The llama 3 herd of models, 2024.
\newblock URL \url{https://arxiv.org/abs/2407.21783}.

\bibitem[Templeton et~al.(2024)Templeton, Conerly, Marcus, Lindsey, Bricken, Chen, Pearce, Citro, Ameisen, Jones, et~al.]{templeton2024scaling}
Adly Templeton, Tom Conerly, Jonathan Marcus, Jack Lindsey, Trenton Bricken, Brian Chen, Adam Pearce, Craig Citro, Emmanuel Ameisen, Andy Jones, et~al.
\newblock Scaling monosemanticity: Extracting interpretable features from claude 3 sonnet.
\newblock Technical report, Anthropic, 2024.

\bibitem[Yang et~al.(2024)Yang, Yang, Zhang, Hui, Zheng, Yu, Li, Liu, Huang, Wei, Lin, Yang, Tu, Zhang, Yang, Yang, Zhou, Lin, Dang, Lu, Bao, Yang, Yu, Li, Xue, Zhang, Zhu, Men, Lin, Li, Xia, Ren, Ren, Fan, Su, Zhang, Wan, Liu, Cui, Zhang, and Qiu]{qwen2.5}
An~Yang, Baosong Yang, Beichen Zhang, Binyuan Hui, Bo~Zheng, Bowen Yu, Chengyuan Li, Dayiheng Liu, Fei Huang, Haoran Wei, Huan Lin, Jian Yang, Jianhong Tu, Jianwei Zhang, Jianxin Yang, Jiaxi Yang, Jingren Zhou, Junyang Lin, Kai Dang, Keming Lu, Keqin Bao, Kexin Yang, Le~Yu, Mei Li, Mingfeng Xue, Pei Zhang, Qin Zhu, Rui Men, Runji Lin, Tianhao Li, Tingyu Xia, Xingzhang Ren, Xuancheng Ren, Yang Fan, Yang Su, Yichang Zhang, Yu~Wan, Yuqiong Liu, Zeyu Cui, Zhenru Zhang, and Zihan Qiu.
\newblock Qwen2.5 technical report.
\newblock \emph{arXiv preprint arXiv:2412.15115}, 2024.

\end{thebibliography}
